# Comparative Analysis of Topic Modeling Techniques on ATSB Text Narratives Using Natural Language Processing

Aziida Nanyonga
School of Engineering and Technology
University of New South Wales
Canberra, Australia
a.nanyonga@unsw.edu.au

Hassan Wasswa
School of Systems and Computing
University of New South Wales
Canberra, Australia
h.wasswa@unsw.edu.au

Ugur Turhan
School of Science
University of New South Wales
Canberra, Australia
u.turhan@unsw.edu.au

Keith Joiner
School of Engineering and Technology
University of New South Wales
Canberra, Australia
k.joiner@unsw.edu.au

Graham Wild
School of Science
University of New South Wales
Canberra, Australia
g.wild@unsw.edu.au

*Abstract*—Improvements in aviation safety analysis call for innovative techniques to extract valuable insights from the abundance of textual data available in accident reports. This paper explores the application of four prominent topic modelling techniques, namely Probabilistic Latent Semantic Analysis (pLSA), Latent Semantic Analysis (LSA), Latent Dirichlet Allocation (LDA), and Non-negative Matrix Factorization (NMF), to dissect aviation incident narratives using the Australian Transport Safety Bureau (ATSB) dataset. The study examines each technique's ability to unveil latent thematic structures within the data, providing safety professionals with a systematic approach to gain actionable insights. Through a comparative analysis, this research not only showcases the potential of these methods in aviation safety but also elucidates their distinct advantages and limitations.

Keywords—Topic Modelling, Aviation Safety, NLP, ATSB

## I. INTRODUCTION

The aviation industry plays a pivotal role in global transportation, ensuring the safe and efficient movement of people and goods across the world. As aviation activities continue to expand, safety remains a paramount concern. To ensure the highest levels of safety, aviation organizations worldwide collect and analyze a wealth of data, particularly incident reports. These reports, submitted to entities like the Australian Transport Safety Bureau (ATSB), contain crucial information about safety incidents, accidents, and near misses [1]. Analyzing and extracting insights from this vast textual data is an essential endeavour, as it can lead to improved safety measures, accident prevention, and enhanced risk assessment within the aviation sector.

The volume of textual data generated by aviation incident reports is immense, making manual analysis both time-consuming and impractical. To address this challenge, automated text analysis methods, known as topic modelling, have emerged as powerful tools. Topic modelling techniques are applied to discover latent thematic structures within textual data, enabling the extraction of relevant information, trends, and patterns [2]–[4].

In this paper, we present a comparative analysis of four leading topic modelling techniques; Probabilistic Latent Semantic Analysis (pLSA), Latent Semantic Analysis (LSA), Latent Dirichlet Allocation (LDA), and Non-negative Matrix Factorization (NMF) as applied to the text narratives found in the ATSB dataset. We explore the potential of Natural Language Processing (NLP) techniques to preprocess the data and facilitate the application of these topic modelling methods.

The primary objective of this study is to evaluate and compare the performance of these topic modelling techniques in terms of their ability to uncover critical insights from aviation incident narratives. Our research aims to contribute to the ongoing efforts to enhance aviation safety and risk assessment through advanced text analysis.

In the following sections of this paper, we will delve into the related work in Section II, and our methodology in Section III, present the experimental result, and engage in a comprehensive discussion of our findings in Section IV. Furthermore, we will explore the implications of our research for the field of aviation safety and suggest potential avenues for future research in Section V.

## II. RELATED WORK

Topic modelling is a critical issue of text analysis and natural language processing, providing a potent means to unveil latent thematic structures within extensive volumes of textual data. In recent years, the application of topic modelling techniques to aviation incident narratives and the utilization of NLP methods in this context have gained significant traction. Traditionally, accident investigation and safety analysis predominantly relied on expert analysis and statistical methods [2], [5], [6]. Nonetheless, these conventional methodologies are hampered by their dependence on the manual examination of accident reports, which is a time-consuming process and is susceptible to human bias [7]. Experts meticulously scrutinized textual narratives, findings, and recommendations to discern recurring patterns and contributing factors. While these methods have yielded valuable insights, they have limitations when dealing with the extensive and intricate datasets that aviation accident reports represent.

Recent advancements in automated text analysis techniques have ushered in a new era for aviation safety research. Researchers have acknowledged the potential of NLP and machine learning to extract actionable insights from textual data [8]–[10]. Text mining and sentiment analysis have been applied to scrutinize and analyze reports such as online forums [11], [12]. These approaches offer scalability and objectivity, mitigating the human bias inherent in traditional analyses.

Topic modelling, a subset of NLP, has garnered significant attention in the realm of aviation safety research. It offers a systematic approach to unveil latent thematic structures within textual data, making it particularly suited for aviation accident reports [13]. Blei et al. [14] introduced LDA, a seminal topic modelling algorithm that has become a cornerstone in this field. LDA has found applications in diverse textual datasets, spanning from news articles and social media content to scientific literature. Its proficiency in revealing underlying topics and relationships has proven invaluable across numerous contexts. Additionally, Lee and Seung introduced NMF, a dimensionality reduction technique applied in text mining and topic modelling. NMF has been utilized in various studies to extract topics from textual data, offering an alternative approach to LDA [15].

The application of topic modelling techniques to aviation incident reports holds substantial promise. pLSA and LDA have been utilized to unveil latent topics within textual narratives [4]. In a study by Luo and Shi, LDA was employed to identify topics related to accident causation and contributing factors within aviation accident reports, effectively showcasing the extraction of meaningful topics [16].

Effective text preprocessing is a vital step in text analysis. The application of NLP techniques, such as text tokenization, stop-word removal, and stemming, has become a standard practice in the field. Research by Alghamdi & Alfalqi underscored the significance of NLP preprocessing in optimizing the performance of topic modelling techniques. Their survey conclusively demonstrated that proper preprocessing significantly enhances the quality of topics extracted from textual data [17].

Latent Semantic Analysis (LSA) has also found applications in aviation safety analysis. Researchers have employed LSA to unveil concealed structures in safety reports, thereby facilitating enhanced risk assessment and safety measures. LSA was deployed to analyze the latent semantic structure of safety reports within the aviation industry, with the intention of improving risk assessment and accident prevention strategies [18].

Non-negative Matrix Factorization (NMF) has gained prominence in text analysis for its capability to provide interpretable and non-negative factorization of document-term matrices. Within the context of aviation safety, NMF has been used to extract topics from incident reports. Li, Shen, and Xu delved into the application of NMF to identify safety topics from incident narratives, effectively demonstrating its capacity to yield meaningful and interpretable topics [19].

Robinson [13] harnessed LDA to extract topics from aviation safety reports and identify emerging safety concerns. Similarly, Ahadh et al. [5] delved into topic modelling to categorize narratives in aviation accident reports, furnishing a structured representation of accident data. [2] made use of text-mining techniques to analyze aviation accident reports, with a focus on identifying significant terms and phrases, thereby establishing the groundwork for the application of computational methods in accident report analysis. Zhong et al. [20] introduced a framework that seamlessly melded text mining and machine learning, enabling the automated classification of accident reports into categories based on contributing factors. Their approach served as a testament to the potential for automating key aspects of accident analysis, consequently enhancing efficiency and consistency.

The work by Rose et al. [21] delivered a compelling contribution by employing structural topic modelling, specifically LDA, to aviation accident reports. Their research not only affirmed the feasibility of employing topic modelling to reveal latent themes within accident narratives but also underscored its potential to automate certain facets of the analysis process, leading to enhanced efficiency. Their research emphasized the critical importance of selecting an appropriate topic modelling technique tailored to specific domains and datasets, accentuating the need for a nuanced approach to topic modelling in aviation safety analysis.

Collectively, these studies underscore the significance of text analysis and topic modelling in the enhancement of aviation safety. They emphasize the importance of adopting advanced techniques, including NLP preprocessing and conducting comparative evaluations of topic modelling methods to extract actionable insights from aviation incident reports. In our research, we aim to build upon these foundations by conducting a comprehensive comparative analysis of four prominent topic modelling techniques; pLSA, LSA, LDA, and NMF applied to the ATSB dataset. Our work contributes to the ongoing efforts to bolster aviation safety and risk assessment through advanced text analysis.

## III. METHODOLOGY

This section outlines the processes and techniques employed in this research to conduct a comparative analysis of four prominent topic modelling techniques: pLSA, LSA, LDA, and NMF. This section provides an in-depth explanation of data collection, preprocessing, and the implementation of each topic modelling method.

### A. Data Collection

Aviation incident/accident investigation reports are collected and published by various organizations such as ATSB, the Aviation Safety Reporting System (ASRS), and the National Transportation Safety Board (NTSB). For this study, the researchers utilized the ATSB aviation incident/accident investigation reports. Depending on the nature of the problem, we considered text narratives that were recorded in Australia for the period of 10 years resulting in a dataset with 53,275 records where the data was sourced directly from the ATSB investigation authorities spanning from 1/01/2013 to 12/31/2022. This resulted in a dataset comprising 50,778 records following data preprocessing and cleaning.

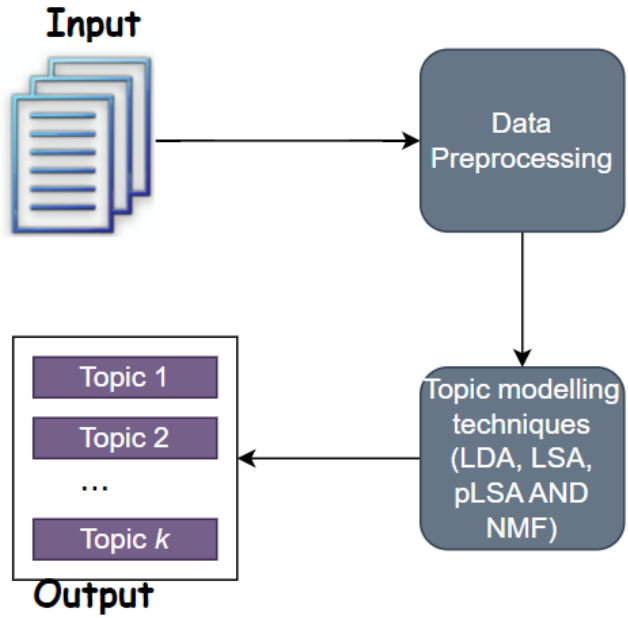

Fig. 1. Methodological framework

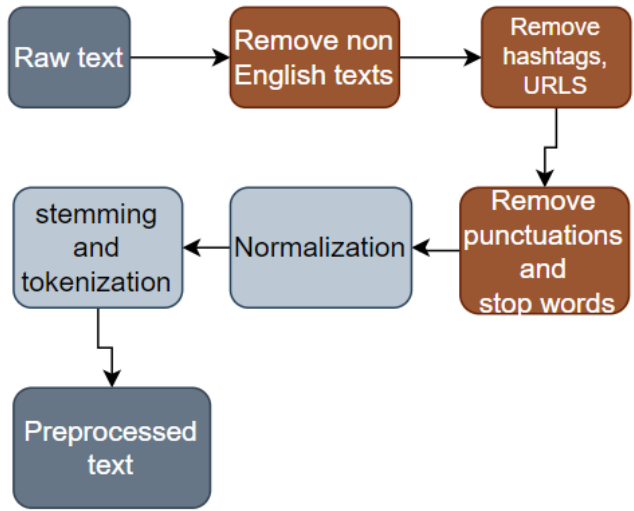

Fig. 2. Text preprocessing

## B. Text Preprocessing

Machine learning models inherently lack the capability to comprehend raw textual data. Our text preprocessing pipeline encompasses several essential stages to enhance data quality and improve model performance. As shown in Fig. 2, these stages include lowercasing, punctuation removal, tokenization, stopword removal, and URL removal. Lowercasing ensures consistency in the text, while punctuation removal streamlines the text for analysis. Tokenization breaks down narratives into individual words, enabling further analysis. Stopword removal eliminates common stopwords, reducing noise, and URL removal ensures web links do not interfere with the analysis.

Once narratives undergo these preprocessing steps, they become ready for feature extraction, a vital transformation that converts textual data into numerical features suitable for machine learning models. For feature extraction, we utilized two distinct techniques: Term Frequency-Inverse Document Frequency (TF-IDF) and Word Embeddings (Word2Vec). TF-IDF quantifies term importance within the narratives, capturing the semantic meaning of words. Word2Vec represents words as dense vectors, allowing models to understand semantic relationships. Additional preprocessing steps included the removal of HTML tags, non-alphanumeric characters, and other irrelevant elements. Consistent lowercase conversion and stopword removal reduced noise, and tokenization created a suitable corpus. Finally, lemmatization reduced words to their base forms, improving topic modelling interpretability.

These comprehensive preprocessing and feature extraction steps ensure the textual data is transformed into a suitable format for subsequent topic modelling, leading to a more robust analysis of aviation accident reports.

## C. Topic Modelling Procedure

After thorough text preprocessing, the next crucial steps involve transforming the preprocessed textual data into numerical features suitable for topic modelling techniques; pLSA, LSA, LDA, and NMF. Each technique was implemented separately, and topics were extracted from the dataset as shown in Fig. 1.

The preprocessed textual data were skillfully transformed into a Document-Term Frequency Matrix. This matrix plays a pivotal role in representing the frequency of each word across all narratives contained in the accident reports. Essentially, it provides a structured numerical representation of the textual data, facilitating the subsequent topic modelling process.

### 1) Latent Dirichlet Allocation (LDA)

For the task of topic modelling, we harnessed the capabilities of LDA. LDA, a probabilistic generative model, is built upon the fundamental assumption that documents represent blends of various topics, and these topics, in turn, comprise assortments of words. LDA's reputation as a stalwart in the realm of topic modelling is well-documented [14]. This model stands out for its proficiency in unveiling the concealed thematic architectures inherent in textual data, rendering it an intuitive selection for our analytical endeavours.

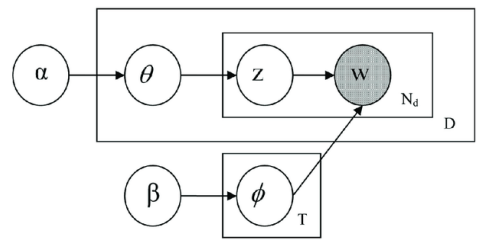

Fig. 3. Shows a probabilistic graphical representation of the LDA model

### 2) Latent Semantic Analysis (LSA)

Latent Semantic Analysis (LSA) is a dimensionality reduction technique that transforms the document-term matrix into a lower-dimensional space. LSA identifies the underlying structure of the text data by uncovering relationships between terms and documents [22]. We used LSA to extract topics from the aviation safety reports, providing insights into the latent themes within the corpus.

### 3) Probabilistic Latent Semantic Analysis (pLSA)

pLSA, short for Probabilistic Latent Semantic Analysis, adopts a probabilistic approach in contrast to

Singular Value Decomposition (SVD) to address the topic modelling problem [23]. The fundamental concept behind pLSA is to establish a probabilistic model featuring latent topics that can generate the observed data in our document-term matrix. More specifically, we seek a model *P (D, W)* that assigns probabilities to each entry in the document-term matrix for any given document *d* and word *w*. In line with the foundational assumptions of topic models, which propose that each document is a blend of various topics, and each topic is composed of a set of words, pLSA introduces a probabilistic twist to these principles:
1. When considering a document d, pLSA assigns topic *z* to that document with the likelihood denoted as *P(z|d)*.
2. When contemplating a topic z, pLSA models the probability of drawing a word w from that topic as *P(w|z)*.

$$P(D,W) = P(D) \sum_Z P(Z|D) P(W|Z)$$

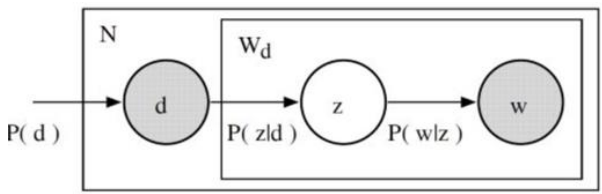

Fig. 4. Shows a representation of the pLSA model adopted from [24]

In a more intuitive sense, the expression on the right-hand side of this equation indicates the likelihood of encountering a particular document. It further considers the distribution of topics within that document and calculates the likelihood of finding a specific word within that document. We have parameters *P(D), P(Z|D),* and *P(W|Z)* in our model. *P(D)* can be directly computed from our corpus data. For *P(Z|D)* and *P(W|Z)*, they are represented as multinomial distributions and can be trained through the expectation-maximization algorithm (EM) [24]. To provide a simplified explanation, EM is a technique used to determine the most probable parameter estimates for a model that relies on unobserved, latent variables (in this context, the topics).

*4) Non-negative Matrix Factorization (NMF).*
Non-negative Matrix Factorization (NMF) is an alternative dimensionality reduction technique within the realm of topic modelling. NMF operates by factorizing the Document-Term Frequency Matrix into two distinct lower-dimensional matrices: one that characterizes topics and the other that represents term distributions [15]. Notably, what distinguishes NMF is its inherent interpretability, making it a valuable tool for extracting meaningful insights from aviation accident reports. The dataset is represented as a *w×d*, matrix, V. where w represents words in each document, d. Fig 4 illustrates a simple mechanism of how NMF breaks V into its constituent components, W and H where W is a *w×t* matrix, H is a *t×d*, matrix, and t represents the distinct topics in V.

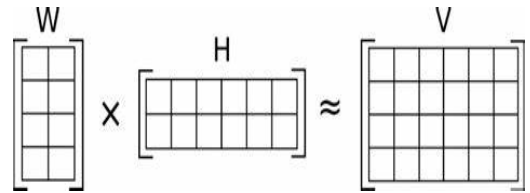

Fig. 5. NMF Model

## IV. RESULTS AND DISCUSSION

This section of the paper provides an overview of the outcomes of the topic modelling analysis conducted using four prominent techniques: pLSA, LSA, LDA, and NMF applied to the ATSB dataset.

### A. Topic Extraction Using Plsa

The application of pLSA to the ATSB dataset resulted in the extraction of ten distinct topics. Each topic is represented by a list of words, and their corresponding probabilities provide insights into the thematic content of aviation incident narratives. Table I presents an overview of the identified topics and their associated words.

TABLE I. TOPIC EXTRACTED BY PLSA AND INTERPRETATION

| Topic No. | Topics | Thematic Keywords |
|---|---|---|
| 0 | Bird Strikes | flight, detected, strike, evidence, birdstrike |
| 1 | Pilot and Aircraft Damage | pilot, damage, minor, helicopter, sustained |
| 2 | Safety Inspection | runway, safety, inspection, retrieved, flying |
| 3 | Engineering and Engine Issues | engine, engineering, cruise, detected, revealed |
| 4 | Cockpit and Descent | cockpit, descent, observed, runway, pilot |
| 5 | Routine Radio Communication | routine, radio, approach, calls, crew |
| 6 | Air Traffic Control (ATC) and Clearance | ATC, clearance, runway, approach, separation |
| 7 | Landing Gear | landing, gear, approach, failed, aircraft |
| 8 | Aircraft Strikes | aircraft, struck, bird, approach, encountered |
| 9 | Takeoff and Aircraft Strikes | takeoff, aircraft, struck, crew, kite |

Table 1 provides an overview of thematic clusters within aviation incident narratives. Each thematic cluster is defined by a list of associated keywords, providing insights into the content of aviation safety reports. These keywords are a representation of the most frequent terms found within each topic.

In addition to the table, we have created thematic visualizations in the form of word clouds. These visual representations offer an at-a-glance view of the most prominent keywords within each thematic cluster. These word clouds visually represent each thematic cluster's most frequently occurring words. They provide a quick overview of the narrative content related to each topic as shown in Fig. 5.

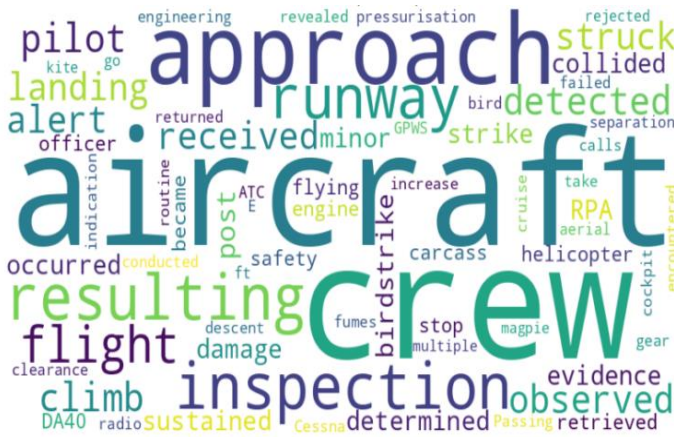

Fig. 6. Word clouds on the pLSA Model

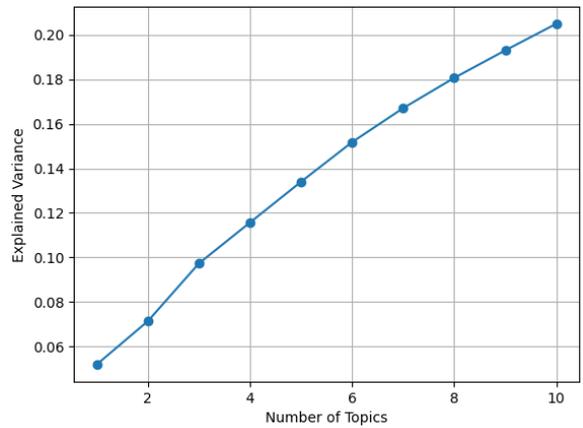

Fig. 7. LSA variance for different topics.

The extraction of these topics through pLSA provides valuable insights into the underlying thematic structures of aviation incident narratives. These topics encompass a wide range of safety-related events, including bird strikes, pilot and aircraft damage, safety inspections, engine issues, cockpit dynamics, routine communication, air traffic control, landing gear problems, aircraft strikes, and takeoff incidents. Understanding the contexts, contributing factors, and commonalities within these topics can guide aviation safety professionals in risk assessment, accident prevention, and safety enhancements.

The pLSA model offers a systematic approach to uncovering latent thematic structures within textual data, making it particularly well-suited for aviation incident reports. These identified topics serve as a foundation for further investigations, enabling researchers and aviation safety organizations to delve deeper into the narratives and extract actionable insights.

### B. Topic Extraction Using LSA

Latent Semantic Analysis (LSA) leverages mathematical techniques to uncover underlying themes within aviation incident narratives. We present an overview of the identified topics, their associated words, and the proportion of variance explained, as well as a thematic word cloud visual representation of these topics.

As part of the analysis, we examined the explained variance with respect to the number of topics as seen in Fig. 7. This visualization helps determine the optimal number of topics to capture the underlying thematic structure of the aviation incident narratives.

The LSA model has successfully unveiled these ten distinct topics from the aviation incident narratives. These topics encompass a wide range of safety-related events and provide a foundation for deeper explorations and analyses. Table II shows the topics and the thematic keywords selected by the LSA model

TABLE II. TOPIC EXTRACTED BY LSA AND INTERPRETATION

| Topic No. | Topics | Thematic Keywords |
|---|---|---|
| 0 | Aircraft and Flight Operations | aircraft, crew, pilot, landing, approach |
| 1 | Crew and Aircraft Inspections | crew, aircraft, pilot, approach, struck |
| 2 | Pilot and Bird Strikes | pilot, struck, flight, engine, bird |
| 3 | Approach and Safety Inspections | approach, landing, inspection, runway, struck |
| 4 | Landing Gear and Flight | landing, gear, approach, aircraft, flight |
| 5 | Runway Inspections and Safety | runway, inspection, crew, pilot, officer |
| 6 | Runway and Flight Occurrences | runway, approach, crew, flight, struck |
| 7 | Engine Issues and Flight Operations | engine, flight, landing, approach, crew, inspection |
| 8 | Takeoff and Helicopter Strikes | take, struck, helicopter, aircraft, bird |
| 9 | Cabin Fumes and Engine Problems | fumes, engine, flight, cabin, detected |

### C. Topic Extraction Using LDA

We present an overview of the identified topics, their associated words, the distribution of topics across the dataset, and a thematic word cloud visualization. Fig. 8 illustrates the top words for each of the ten topics. This visualization provides a snapshot of the most significant terms that define each topic.

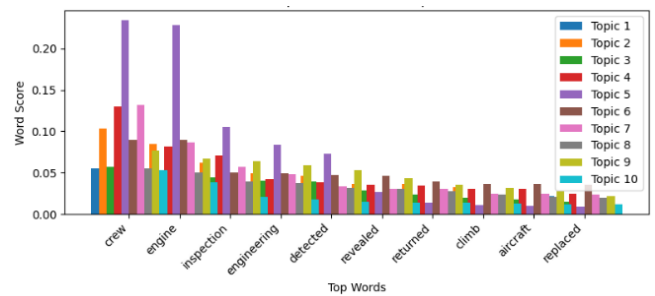

Fig. 8. Top words chosen by each topic on the LDA Model

Another visual representation showcases the distribution of these topics across the aviation incident narratives. As shown in Fig. 9, Topics 1 and 4 emerge as the most significant themes in the dataset, offering valuable insights into the narrative content.

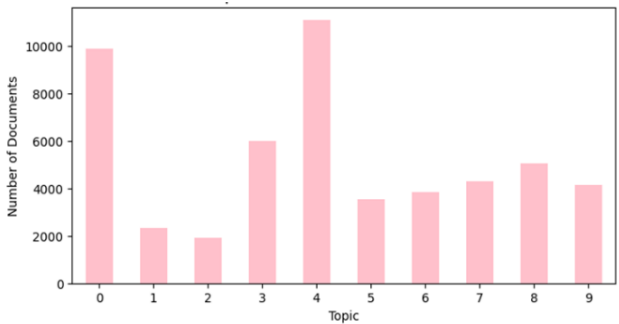

Fig. 9. Topic distribution for LDA Model

Thematic word clouds offer a visual summary of each topic shown in Fig. 10. By highlighting the most frequently occurring words within a topic, they provide an at-a-glance understanding of the thematic content of the narratives.

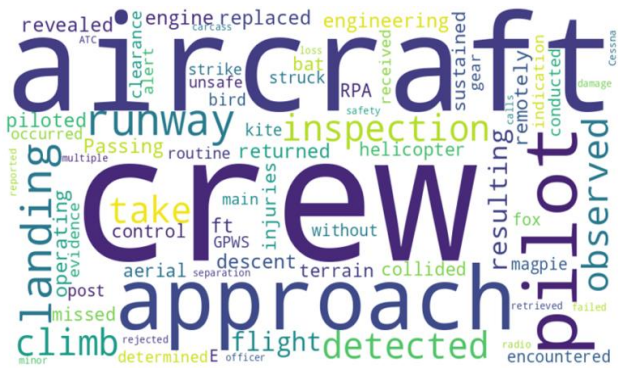

Fig. 10. Word Cloud of Topics on LDA Model

LDA successfully uncovers thematic structures within aviation incident narratives. These topics encompass a wide range of safety-related events, from engine issues and flight operations to pilot incidents and bird strikes. These insights serve as a valuable resource for further analysis and enhancing aviation safety.

### D. Topic Extraction Using NMF

For NMF successfully identified ten topics within the aviation incident narratives, each represented by a set of associated keywords. Thematic word clouds offer a visual summary of each topic. These topics span a wide range of safety-related events, encompassing engine and aircraft operations, pilot incidents, bird strikes, post-flight inspections, landing gear, runway safety, aircraft damage, and more as shown in Fig. 11. These insights serve as valuable resources for further analysis and enhancing aviation safety.

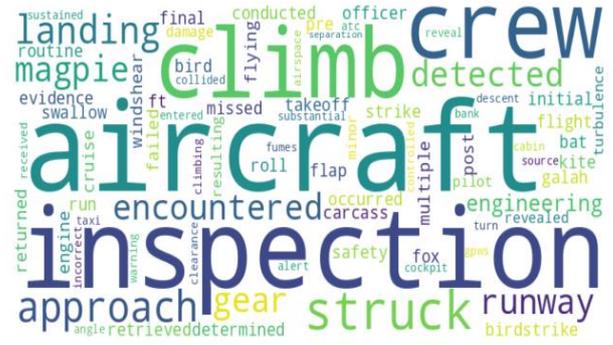

Fig. 11. Word Cloud of Topics on NMF Model

Among the ten identified topics, topics 1, 4, 7, and 8 were selected as the most significant for further analysis and investigation due to their prominence and relevance to aviation safety as shown in Fig. 12.

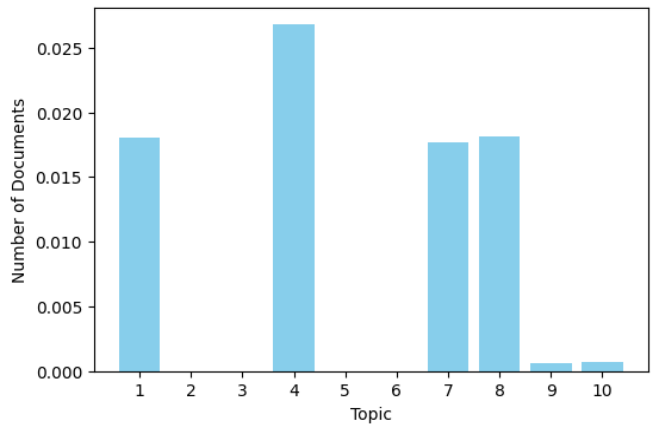

Fig. 12. Topic distribution for NMF Model

NMF successfully reveals the thematic structures within the aviation incident narratives. These topics span a wide range of safety-related events, encompassing engine and aircraft operations, pilot incidents, bird strikes, post-flight inspections, landing gear, runway safety, aircraft damage, and more. These insights serve as a valuable resource for further analysis and enhancing aviation safety.

### E. Comparative Analysis of Topic Modelling Techniques

We present a comparative analysis of the three prominent topic modelling techniques used in our study: Probabilistic pLSA, LSA, and LDA, as well as the NMF. The goal is to highlight their respective strengths, weaknesses, and contributions to aviation safety analysis using the ATSB dataset.

Each topic modelling technique offers unique advantages and limitations. pLSA excels in comprehensive topic discovery and interpretability but may face data-dependent challenges and scalability issues. LSA focuses on dimensionality and noise reduction, enhancing information retrieval, but is sensitive to preprocessing choices. LDA provides topic distributions and document-topic relations, though its complexity, sensitivity to hyperparameters, and interpretational challenges can be drawbacks. NMF enforces non-negativity constraints, ensuring interpretable topics, and scalability, but it may encounter issues with sparse data and is restricted to positive data as summarized in Table III.

Our comparative analysis reveals that each of the four topic modelling techniques: pLSA, LSA, LDA, and NMF, offers unique advantages and challenges. The choice of which technique to employ depends on the specific goals and characteristics of the dataset. While pLSA excels in comprehensive topic discovery and interpretability, LSA focuses on dimensionality reduction and noise reduction. LDA provides a probabilistic framework for topic modelling, and NMF enforces non-negativity constraints for interpretable results.

TABLE III. COMPARATIVE ANALYSIS OF TOPIC MODELLING TECHNIQUES.

| Topic Modelling Technique | Strengths | Weaknesses |
|---|---|---|
| pLSA | Comprehensive Topic Discovery | Data, Dependent |
| | Interpretability | Limited Scalability |
| | Actionable Insights | Topic Overlaps |
| LSA | Dimensionality Reduction | Limited Latent Structure |
| | Noise Reduction | Dependent on Preprocessing |
| | Improved Information Retrieval | Complexity |
| LDA | Generative Model | Sensitivity to Hyperparameters |
| | Topic Distribution | Difficulty in Topic Interpretation |
| | Document-Topic Relations | Complexity |
| NMF | Non-Negativity Constraint | Limited to Positive Data |
| | Scalability | Difficulty in Handling Sparse Data |
| | Interpretable Topics | Manual Selection of Topics |

## V. CONCLUSION

In this study, we delved into the realm of aviation safety analysis, focusing on four distinct topic modelling techniques; pLSA, LSA, LDA, and NMF applied to aviation incident narratives from the ATSB dataset. These techniques have exhibited their capacity to uncover hidden thematic structures within textual data, providing valuable insights into aviation safety reports. Despite their unique characteristics, each method contributes to the automation of key aspects of accident analysis, thereby mitigating human bias and enhancing safety assessment. Our comparative analysis has unveiled the varying performance of these techniques, enabling aviation safety organizations to make informed choices when selecting the most suitable method for their specific requirements. As we move forward in aviation safety analysis, the integration of natural language processing and topic modelling will play a pivotal role, ultimately contributing to safer skies and more informed risk assessment.

Future work in the field of aviation safety analysis should focus on several key areas. Firstly, the integration of more advanced natural language processing techniques, such as deep learning and recurrent neural networks, can enhance the accuracy and depth of insights derived from accident narratives. Additionally, a more extensive exploration of ensemble methods that combine the strengths of different topic modelling techniques may provide a holistic view of aviation incident data. Furthermore, the development of domain-specific topic modelling methods tailored to aviation safety narratives can improve the precision and interpretability of results. Finally, the adoption of real-time incident data streams and the development of predictive modelling tools for proactive risk assessment can further contribute to the overarching goal of enhancing aviation safety.

REFERENCES

[1] A. Somerville, T. Lynar, and G. Wild, "The nature and costs of civil aviation flight training safety occurrences," *Transportation Engineering*, vol. 12, Jun. 2023, doi: 10.1016/j.treng.2023.100182.

[2] K. D. Kuhn, "Using structural topic modelling to identify latent topics and trends in aviation incident reports," *Transp Res Part C Emerg Technol*, vol. 87, pp. 105–122, Feb. 2018, doi: 10.1016/j.trc.2017.12.018.

[3] C. Paradis, R. Kazman, M. D. Davies, and B. L. Hooey, "Augmenting topic finding in the nasa aviation safety reporting system using topic modelling," in *AIAA Scitech 2021 Forum*, American Institute of Aeronautics and Astronautics Inc, AIAA, 2021, pp. 1–16. doi: 10.2514/6.2021-1981.

[4] K. Stevens, P. Kegelmeyer, D. Andrzejewski, and D. Buttler, "Exploring Topic Coherence over many models and many topics," in *In Proceedings of the 2012 joint conference on empirical methods in natural language processing and computational natural language learning*, Association for Computational Linguistics, 2012, pp. 952–961. [Online]. Available: http://mallet.cs.umass.edu/

[5] A. Ahadh, G. V. Binish, and R. Srinivasan, "Text mining of accident reports using semi-supervised keyword extraction and topic modelling," *Process Safety and Environmental Protection*, vol. 155, pp. 455–465, Nov. 2021, doi: 10.1016/j.psep.2021.09.022.

[6] V. de Vries, "Classification of aviation safety reports using machine learning," In 2020 International Conference on Artificial Intelligence and Data Analytics for Air Transportation (AIDA-AT), pp. 1-6. IEEE, 2020., Feb. 2020, pp. 1–6.

[7] R. K. Dismukes, B. A. , Berman, and L. Loukopoulos, *The limits of expertise: Rethinking pilot error and the causes of airline accidents*. Routledge, 2017.

[8] A. Miyamoto, M. V. Bendarkar, and D. N. Mavris, "Natural Language Processing of Aviation Safety Reports to Identify Inefficient Operational Patterns," *Aerospace*, vol. 9, no. 8, Aug. 2022, doi: 10.3390/aerospace9080450.

[9] A. Nanyonga, H. Wasswa, U. Turhan, O. Molloy, and G. Wild, "Sequential Classification of Aviation Safety Occurrences with Natural Language Processing," in *In AIAA AVIATION 2023 Forum*, American Institute of Aeronautics and Astronautics

(AIAA), Jun. 2023, p. 4325. doi: 10.2514/6.2023-4325.
[10] A. Nanyonga, H. Wasswa, O. Molloy, U. Turhan, and G. Wild, "Natural Language Processing and Deep Learning Models to Classify Phase of Flight in Aviation Safety Occurrences," in *2023 IEEE Region 10 Symposium (TENSYMP)*, IEEE, Sep. 2023, pp. 1–6. doi: 10.1109/TENSYMP55890.2023.10223666.
[11] N. Li and D. D. Wu, "Using text mining and sentiment analysis for online forums hotspot detection and forecast," *Decis Support Syst*, vol. 48, no. 2, pp. 354–368, Jan. 2010, doi: 10.1016/j.dss.2009.09.003.
[12] S. Wakade, C. Shekar, K. J. Liszka, and C.-C. Chan, "Text Mining for Sentiment Analysis of Twitter Data." [Online]. Available: http://jaiku.com
[13] S. D. Robinson, "Temporal topic modelling applied to aviation safety reports: A subject matter expert review," *Saf Sci*, vol. 116, pp. 275–286, Jul. 2019, doi: 10.1016/j.ssci.2019.03.014.
[14] D. M. Blei, A. Y. Ng, and J. B. Edu, "Latent Dirichlet Allocation Michael I. Jordan," 2003.
[15] D. D. Lee and H. S. Seung, "Learning the parts of objects by non-negative matrix factorization.," *Nature*, vol. 401, no. 6755, pp. 788–791, 1999.
[16] Y. Luo and H. Shi, "Using lda2vec topic modelling to identify latent topics in aviation safety reports.," in *In 2019 IEEE/ACIS 18th International Conference on Computer and Information Science (ICIS) IEEE.*, 2019, pp. 518–523.
[17] R. Alghamdi and K. Alfalqi, "A survey of topic modelling in text mining.," *Int. J. Adv. Comput. Sci. Appl.(IJACSA)*, vol. 6, no. 1, 2015, [Online]. Available: www.ijacsa.thesai.org
[18] S. D. Robinson, "Multi-label classification of contributing causal factors in self-reported safety narratives," *Safety*, vol. 4, no. 3, 2018, doi: 10.3390/safety4030030.
[19] M. W. Berry, Nicolas Gillis, and Franc̦ois Glineur, "Document Classification Using Nonnegative Matrix Factorization and Underapproximation.," Taipei, Taiwan: 2009 IEEE International Symposium on Circuits and Systems: circuits and systems for human centric smart living technologies, conference program, Taipei International Convention Center, , May 2009, p. 3209.
[20] B. Zhong, X. Pan, P. E. D. Love, J. Sun, and C. Tao, "Hazard analysis: A deep learning and text mining framework for accident prevention," *Advanced Engineering Informatics*, vol. 46, Oct. 2020, doi: 10.1016/j.aei.2020.101152.
[21] R. L. Rose, T. G. Puranik, D. N. Mavris, and A. H. Rao, "Application of structural topic modelling to aviation safety data," *Reliab Eng Syst Saf*, vol. 224, Aug. 2022, doi: 10.1016/j.ress.2022.108522.
[22] T. K. Landauer, P. W. Foltz, and D. Laham, "An introduction to latent semantic analysis," *Discourse Process*, vol. 25, no. 2–3, pp. 259–284, Jan. 1998, doi: 10.1080/01638539809545028.
[23] T. Hofmann, "Probabilistic Latent Semantic Analysis," *arXiv preprint arXiv:*, pp. 1301–6705, 2013.
[24] N. C. Albanese, "Topic Modelling with LSA, pLSA, LDA, NMF, BERTopic, Top2Vec: a Comparison A comparison between different topic modelling strategies including practical Python examples.," 2022.